\documentclass[lettersize,journal]{IEEEtran}
\usepackage{amsmath,amsfonts}
\usepackage{algorithmic}
\usepackage{array}
\usepackage[caption=false,font=small,labelfont=rm,textfont=rm]{subfig}
\usepackage{textcomp}
\usepackage{stfloats}
\usepackage{url}
\usepackage{verbatim}
\usepackage{graphicx}
\def\BibTeX{{\rm B\kern-.05em{\sc i\kern-.025em b}\kern-.08em
		T\kern-.1667em\lower.7ex\hbox{E}\kern-.125emX}}
\usepackage{balance}
\usepackage[utf8]{inputenc}
\usepackage[ruled,vlined]{algorithm2e}
\usepackage{hyperref}

\hypersetup{
	colorlinks = true, 
	linkcolor = blue,
	urlcolor = blue,
	citecolor = blue
}
\begin{document}
	\title{Reconfigurable Intelligent Surface Aided Vehicular Edge Computing: Joint Phase-shift Optimization and Multi-User Power Allocation}
	\author{Kangwei Qi, Qiong Wu,~\IEEEmembership{Senior Member,~IEEE}, Pingyi Fan,~\IEEEmembership{Senior Member,~IEEE}, \\Nan Cheng,~\IEEEmembership{Senior Member,~IEEE}, Wen Chen,~\IEEEmembership{Senior Member,~IEEE}, and Khaled B. Letaief,~\IEEEmembership{Fellow,~IEEE}
		
		\thanks{This work was supported in part by the National Natural Science Foundation of China under Grant 61701197 and Grant 62071296; in part by the National Key Research and Development Program of China under Grant 2021YFA1000500(4); in part by the National Key Project under Grant 2020YFB1807700; in part by Shanghai Kewei under Grant 22JC1404000; in part by the Research Grants Council under the Areas of Excellence Scheme under Grant AoE/E-601/22-R; and in part by the 111 Project under Grant B23008. (Corresponding author: Qiong Wu.)
			
		Kangwei Qi, Qiong Wu are with the School of Internet of Things Engineering, Jiangnan University, Wuxi 214122, China. (e-mail: kangweiqi@stu.jiangnan.edu.cn, qiongwu@jiangnan.edu.cn).
			
		Pingyi Fan is with the Department of Electronic Engineering, Beijing National Research Center for Information Science and Technology, Tsinghua University, Beijing 100084, China (e-mail: fpy@tsinghua.edu.cn).
			
		Nan Cheng is with the State Key Lab. of ISN and School of Telecommunications Engineering, Xidian University, Xi’an 710071, China (e-mail: dr.nan.cheng@ieee.org).
			
		Wen Chen is with the Department of Electronic Engineering, Shanghai Jiao Tong University, Shanghai 200240, China (e-mail: wenchen@sjtu.edu.cn).
			
		Khaled B. Letaief is with the Department of Electrical and Computer Engineering, Hong Kong University of Science and Technology (HKUST), Hong Kong (email:eekhaled@ust.hk).
			}
		}
	
	
	\maketitle
	
	\begin{abstract}
		Vehicular edge computing (VEC) is an emerging technology with significant potential in the field of internet of vehicles (IoV), enabling vehicles to perform intensive computational tasks locally or offload them to nearby edge devices. However, the quality of communication links may be severely deteriorated due to obstacles such as buildings, impeding the offloading process. To address this challenge, we introduce the use of reconfigurable intelligent surface (RIS), which provide alternative communication pathways to assist vehicle communication. By dynamically adjusting the phase-shift of the RIS, the performance of VEC systems can be substantially improved. In this work, we consider a RIS-assisted VEC system, and design an optimal scheme for local execution power, offloading power, and RIS phase-shift, where random task arrivals and channel variations are taken into account. To address the scheme, we propose an innovative deep reinforcement learning (DRL) framework that combines the deep deterministic policy gradient (DDPG) algorithm for optimizing RIS phase-shift coefficients and the multi-agent DDPG (MADDPG) algorithm for optimizing the power allocation of vehicle user (VU). Simulation results show that our proposed scheme outperforms the traditional centralized DDPG, twin delayed deep deterministic policy gradient (TD3) and some typical stochastic schemes.
	\end{abstract}
	

	\section{Introduction}
	\IEEEPARstart{W}{ith} the rapid development of internet of vehicles (IoV), vehicular edge computing (VEC), as a cutting-edge technology, is becoming an important solution to support real-time vehicle computing \cite{r101, cheng102, e6}. Vehicle devices are usually equipped with limited processors and memory, which limits their capability to process complex algorithms and large data \cite{r102, e8}. For example, high-precision real-time positioning services or driver assistance systems that are relied on real-time video analysis require significant computational resources, which may be beyond the capabilities of commonly deployed in-vehicle computing resources \cite{r99, cheng103, cheng107}. However, VEC allows vehicles to offload tasks to edge servers with abundant computing and storage resources so that various functions of the vehicle can be achieved \cite{cheng101, e4, e5}.
	
	Despite extensive research in VEC, challenges remain, particularu in urban environments where communication between vehicle users (VUs) and base stations (BSs) can be obstructed by buildings and other obstacles \cite{e1, e2, e3}. These factors lead to attenuation and multi-path effects on the wireless signals, which in turn affect the stability of the communication links and the transmission rate \cite{r105, r106}. In addition, communication between vehicles and infrastructure needs to be supported by sufficient bandwidth, especially for some applications requiring real-time updates, such as traffic flow monitoring or cooperative driving between vehicles \cite{wu1001, wu1002, e7}. In high-density urban areas or during peak traffic hours, bandwidth may be limited, which can affect the timely transmission and processing of data for each vehicle \cite{wu1005}.
	
	Recently, reconfigurable intelligent surface (RIS) technology has demonstrated significant advantages in enhancing network quality by providing additional communication pathways \cite{wu1003, wu1004, r107}. RIS can dynamically adjust phase-shift to improve communication link quality, thereby enabling better task offloading to edge computing servers \cite{r108, r109}. Therefore, the integration of RIS into VEC systems is a research avenue that deserves to be explored extensively.
	
	However, optimizing the phase-shift of RIS is constrained by hardware conditions, which restrict the phase-shift to a finite set of values. In \cite{r110}, He \emph{et al.} have tackled the RIS phase-shift optimization problem using algorithms like Block Coordinate Descent (BCD), Alternating Optimization (AO), and Semi-definite Relaxation (SDR). However, these algorithms above have a high time complexity and make the RIS phase-shift coefficients adjustment be difficult in time \cite{wu101}. In parallel, artificial intelligence (AI) has been applied to various engineering fields after rapid development \cite{cheng104, e9, cheng105, cheng106, new1, new2}. Deep Reinforcement Learning (DRL), as a type of AI, integrates deep learning (DL) and reinforcement learning (RL) methods, and becomes a promising technique that learns through the interaction of an intelligent body with its environment \cite{new3, new4, wu102, liu101, liu102, liu103}. It can effectively and quickly solve some non-convex problems with high complexity.
	
	There are a number of challenges to leveraging RIS in vehicular communication environments. Firstly, the scenarios are complex and multiple VUs may exist, which can put pressure on training. Second, due to the high-speed mobility of vehicles, their locations are constantly changing, so the distances between the RIS and the VU, and between the VU and the BS, change over time, which affects the quality of the link between them \cite{r111}. Thirdly, the task arrival of each VU is random in practice. Fourthly, phase-shift optimization of RIS is complex, since RIS and VU are not the same type of agents, they may have different actions, That means we have to overcome the phase-shift optimization training of RIS first. Fifth, we should not only consider the power allocation problem of each VU, but also consider the buffer length at the same time. Sixth, how to combine the RIS phase-shift coefficient optimization with the VU power allocation problem is a challenging problem.
	
	The joint optimization of RIS phase-shift and power allocation for multiple VUs is too complex, and centralized algorithms can not work effectively in such dynamic settings. In order to solve the problems mentioned above, we propose a novel power allocation and phase-shift optimization scheme for RIS-assisted VEC system. By using the modified multi-agent deep deterministic policy gradient (MADDPG) algorithm for VU power allocation and the deep deterministic policy gradient (DDPG) algorithm for RIS phase-shift optimization. The contributions of
	this work can be summarized as follows\footnote{The source code has been released at: https://github.com/qiongwu86/DDPG-RIS-MADDPG-POWER}:
	\begin{itemize}
		\item[1)]A system model is proposed that enhances VEC communication using RIS technology. In this model, as the vehicle moves, communication links are impacted by obstacles, leading to reduced link quality. To address this, the VU leverages RIS to improve the quality of communication links and ensures efficient edge computing by offloading tasks to edge servers.
		 
		\item[2)]We investigate the joint optimization of RIS phase-shift coefficients and VU power allocation in RIS-assisted VEC systems. This problem is formulated with the objective of minimizing both power consumption and buffer length for the VUs. However, due to various constraints, it presents as a challenging mixed-integer non-convex problem.
		 
		Since the offloading power and the local execution power are vehicle actions and the phase shift coefficients are RIS actions, the vehicle and the RIS are not the same type of agent. So, we decompose the optimization problem into two sub-problems, firstly, RIS is used as an auxiliary tool which aims to maximize the information transmission rate of the VUs by adjusting the phase-shift coefficients, so we first optimize the phase-shift coefficients of RIS by a centralized DDPG algorithm trained to obtain the phase-shift optimization model. Secondly, for the power allocation of VUs, we consider multiple users and use the modified MADDPG algorithm to solve the power allocation problem where the obtained phase-shift coefficients of RIS in previous phase is adopted as the given part.
		 
		\item[4)]Extensive simulations are used to demonstrate that our proposed scheme significantly outperforms traditional centralized DDPG and twin delayed deep deterministic policy gradient (TD3) schemes as well as some typical stochastic schemes in terms of power consumption and buffer length.
	\end{itemize}
	
	The rest of the paper is organized as follows: Section II reviews related work on VEC and RIS. Section III describes the system model and problem formulation. Section IV and V presents the proposed optimization framework and algorithms. Section VI provides simulation results and analysis. Finally, Section VII concludes the paper.
	
	\section{Related Work}
	In this section, we first review related work on DRL-based offloading schemes for mobile edge computing (MEC), and then review existing work on RIS-assisted VEC offloading.
	\subsection{DRL-based Offloading Scheme for MEC}
	In \cite{r112}, Liu \emph{et al.} explored vehicle-assisted MEC networks and proposed a DRL-based offloading scheme to optimize task offloading and resource allocation.
	In \cite{r113}, Huang \emph{et al.} proposed a DRL-based online offloading framework (DROO) for solving the task offloading problem in wireless-powered MEC networks. A expandable solution is achieved through deep Q-networks (DQN), which obviates the need to solve joint optimization problems and significantly reduces the computational complexity.
	In \cite{R1}, Zhan \emph{et al.} explored the task offloading problem in VEC networks and proposed an offloading scheduling method based DQN to optimize task scheduling and resource allocation.
	In \cite{r114}, Zheng \emph{et al.} combined wireless power transfer (WPT) and multi-access edge computing (MAEC) to overcome energy and computational limitations in wireless devices. They proposed a DRL-based algorithm to minimize total computational delay (TCD) in MEC networks by optimizing offloading decisions, WPT duration, and transmission duration, achieving near-optimal performance with low computational complexity in fast-fading channels.
	In \cite{R2}, Alam \emph{et al.} addressed computational offloading in high-mobility IoV environments, ensuring latency, energy, and cost requirements. They proposed a cooperative three-tier decentralized vehicle assisted MAEC network and applied the multi-agent DRL based Hungarian algorithm (MADRLHA) to optimize dynamic task offloading, enhancing network performance and robustness.
	In \cite{R3}, Liu \emph{et al.} examined vehicular task offloading systems using DRL techniques, focusing on edge cloud (EC) and vehicular cloud (VC) computing. They categorized existing solutions involving MEC servers, nearby vehicles, and hybrid MEC (HMEC), summarizing lessons learned and outlining challenges for future research.
	In \cite{R4}, Zhu \emph{et al.} proposed a decentralized DRL framework to optimize power allocation in multi-input multi-output and
	non-orthogonal multiple access (MIMO-NOMA) VEC systems, addressing uncertainties in channel conditions and stochastic task arrivals.
	In \cite{R5}, Tang \emph{et al.} addressed the challenge of dynamic and efficient offloading decisions in VEC environments by proposing a dynamic framing offloading algorithm based on double DQN. It aims to minimize total delay and waiting time for compute-intensive tasks from multiple moving vehicles by optimizing offloading decisions at the subtask level in a frame-based model.
	In \cite{R6}, Liu \emph{et al.} explored collaborative task computing and dynamic resource allocation in VEC using an asynchronous DRL algorithm. It aims to optimize task offloading and resource management across vehicles, edge servers, and cloud environments.
	In \cite{R7}, Ju \emph{et al.} proposed NOMA-assisted secure offloading in VEC networks. It optimizes transmit power, resource allocation, and jammer vehicle selection to minimize energy consumption while meeting computation delay constraints. Using an asynchronous advantage actor-critic (A3C) learning algorithm, the energy-efficiency secure offloading (EESO) scheme adapts dynamically to VEC environments, improving energy efficiency and ensuring security with low computation delay.
	In \cite{R8}, Wu \emph{et al.} presented a 5G-based vehicle-aware MAEC network, and proposed a DRL-based joint computation offloading and task migration optimization algorithm to optimize computation offloading and task migration.
	
	\subsection{RIS-assisted Offloading Scheme for MEC}
	In \cite{r115}, Zhang \emph{et al.} introduced large-scale model techniques in IoV to tackle coverage challenges in dynamic wireless networks. They proposed a hybrid beamforming method using DRL to optimize beamforming with RIS and BS, demonstrating improved system throughput and coverage.
	In \cite{r116}, Zhang \emph{et al.} explored NOMA techniques with RIS in IoV to enhance signal transmission. They proposed power allocation schemes optimizing channel gain and data rate by adjusting RIS reflector units, enhancing system capacity and ensuring quality of service (QoS).
	In \cite{r117}, Muhammad \emph{et al.} investigated zero-energy RIS (ze-RIS) for VEC in 6G networks. They proposed a DRL-driven RIS-assisted energy-efficient task offloading (DREEO) scheme to save energy and time through power and offloading mechanism control.
	In \cite{r118}, Chen \emph{et al.} introduced MEC and RIS in IoT for low-latency task offloading. They developed an MEC-NOMA framework and RIS-assisted channel allocation algorithms to optimize offloading delays and transmission rates.
	In \cite{r119}, Mao \emph{et al.} applied RIS techniques in wirelessly powered MEC networks to improve energy efficiency and task offloading. They proposed an iterative algorithm to optimize upstream and downstream beamforming, transmission power, RIS time allocation, and IoT device computation frequency for maximizing total computation bits.
	In \cite{R9}, Dai \emph{et al.} integrated RIS with edge computing in 6G networks to enhance wireless communication efficiency and support low-latency applications. It introduces a DRL-based computation offloading scheme to minimize mobile device offloading latency, demonstrating improved data rates and reduced task execution times through RIS-aided enhancements.
	In \cite{R10}, Hazarika \emph{et al.} investigated an IoV network integrating MAEC servers at BSs with RISs for enhanced uplink and downlink transmissions. It employs a MA-DRL framework using Markov game theory to optimize task offloading decisions, improving network utility and communication quality significantly over existing methods.
	In \cite{R11}, Xu \emph{et al.} explored the integration of RIS into MEC networks, focusing on maximizing the sum computation rate. It proposes a DRL-based approach, specifically employing the TD3 algorithm, to optimize RIS phase-shift and energy partition strategies for IoT devices.
	In \cite{R12}, Ning \emph{et al.} developed a DRL framework to jointly optimize computational edge servers, RIS deployment, and beamforming matrices in VEC networks, aiming to maximize the weighted sum throughput of VU equipment while constraining latency within defined thresholds.
	In \cite{R13}, Xu \emph{et al.} studied MEC networks with RIS support to maximize the total computation rate by optimizing the energy allocation strategy for RIS phase-shift and IoT devices, and then solve it by the TD3 algorithm.
	In \cite{R14}, Bin \emph{et al.} focused on optimizing energy consumption in MEC systems using RIS, especially in the context of mobile user equipment (UEs). The goal is to minimize the energy consumption by jointly optimizing the discrete phases of the RIS, the transmit power of the UEs, the computational resource allocation, and the task offloading policy. To cope with this complex problem with time-series decisions and uncertainties, a Soft Actor-Critic (SAC) algorithm based on deep reinforcement learning is proposed.
	In \cite{R15}, a RIS-assisted MEC system is investigated, aiming to minimize the energy consumption of hybrid access point (HAP) by optimizing the offloading strategy of the user, the active beamforming of the hybrid access point and the passive beamforming of the IRS. An optimization-driven hierarchical deep deterministic policy gradient (OH-DDPG) framework is proposed to learn to adjust the RIS's policy via external DDPG and improve the learning efficiency in multi-agent scenarios.
	
	Reviewing the above related studies, we find that although many advances have been made in RIS-assisted VEC, comprehensive studies that simultaneously consider offloading power, local execution power, and phase-shift control are still relatively lacking. This is mainly due to the existing challenges faced by RIS-assisted VEC. In this paper, this is the first time that the integration of power allocation and phase-shift control in the RIS-assisted VEC are decomposed into two sub-problems and solved by the DRL method.
	\section{System Model and Problem Formulation}\label{System}
	\begin{figure}[t]
		\centering
		\includegraphics[width=3.4in]{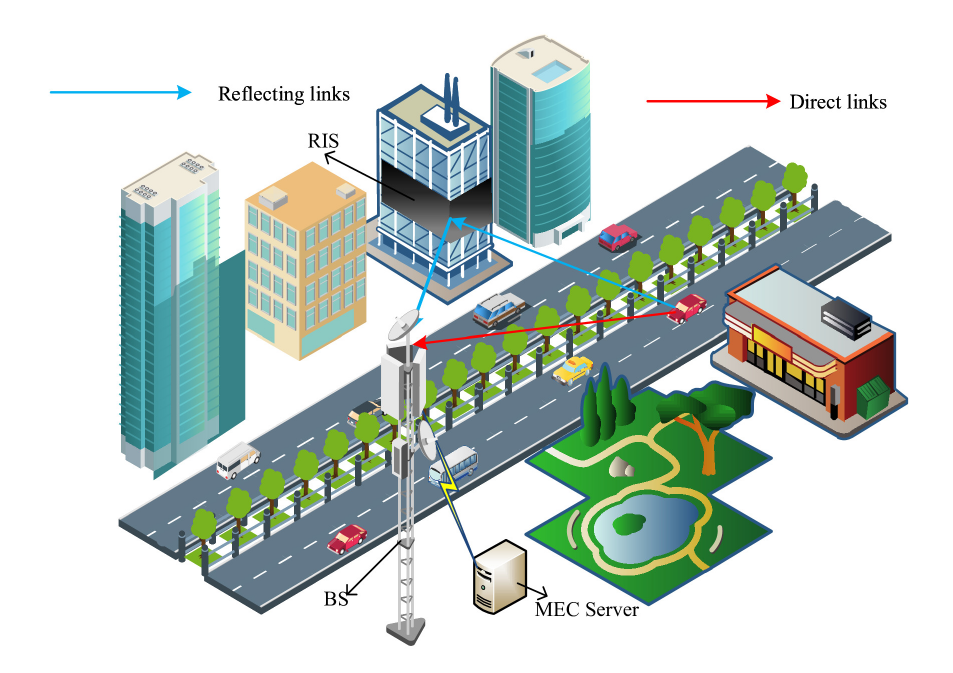}
		\caption{RIS aided vehicular edge computing}
		\label{fig1}
	\end{figure}
	\subsection{Scenario}
	As shown in Fig. \ref{fig1}, we consider a RIS-assisted VEC network with multiple VUs, where the BS is with $M$ antennas and a VEC server, and the VU has a single antenna. Since the VU has limited computational resources, it can offload some tasks to the VEC server for processing. When the VU's link is blocked, the quality of its communication link with the BS is degraded, in which case the RIS with $N$ reflective elements can help the VU offload tasks to the edge device (It assumes that two links existing simultaneously, i.e., the VU communicates with the BS via the RIS and the VU communicates directly with the BS). Let us consider an urban road environment where the RIS is deployed on a building on one side of the road, where there are $K$ VUs, represented in the set as 
	$\mathcal{K} = \left\{{1,2, \cdots ,K} \right\}$. The VUs have the flexibility to offload tasks for processing or local execution, and in addition, we divide the task processing into $T$ equal discrete time slots $\Delta t$, denoted by $\mathcal{T} = \left\{ {1,2, \cdots ,T} \right\}$. At each time slot $t$ in $T$, the VU randomly generates a task whose arrival follows a Poisson distribution with an arrival rate of $\eta$. Note that we consider quasi-static scenarios, i.e., the channel conditions remain constant within a time slot but may change between neighboring time slots.
	
	\subsection{Queue Model}	
	For the $k$th VU, we assume that its task arrival rate at time slot $t$ is 
	${\eta _k}(t)$, which follows a Poisson distribution, so one can calculate the tasks of the $k$th VU during time slot $t$ as 
	\begin{equation}\label{eq1}
		{a_k}(t) = {\eta _k}(t) \times \Delta t,
	\end{equation}
	the arriving tasks will be saved in the buffer, and processed at time slot $t+1$, thus the buffer length of the $k$th VU at time slot $t+1$ is given by
	\begin{equation}\label{eq2}
		{q_k}(t + 1) = {\left[ {{q_k}(t) - {q_{k,o}}(t) - {q_{k,l}}(t)} \right]^ + } + {a_k}(t),
	\end{equation}
	where ${q_{k,o}}(t)$ and ${q_{k,l}}(t)$ denote the amount of tasks processed by task offloading and locally, respectively. Note that ${\left[ x \right]^ + } = \max (0,x)$. In addition, in the process of information transmission, the latency is related to the length of the buffer. For each time slot of buffer, if the processing of the remaining tasks is too long, it may cause a large backlog of data in other parts of the network, which in turn leads to an increase in the overall system latency.
	
	\subsection{Offloading Model}
	In the proposed model, it is assumed that the location of the $k$th VU is $(x_k^t,y_k^t,z_k^t)$ at time slot $t$, and the coordinates of the BS and the RIS are $(B{S_x},B{S_y},B{S_z})$ and $(RI{S_x},RI{S_y},RI{S_z})$, respectively. Therefore, the channel gain between the $\mathit{k}$th VU and BS at slot $\mathit{t}$ is 
	\begin{equation}\label{eq3}
		{\bf{h}}_{k, b}^t = \sqrt {\rho {{(d_{k, b}^t)}^{ -\alpha_{k, b}  }}} {\bf{g}}_{k, b}^t,
	\end{equation}
	where $\rho $ is the path loss constant for the reference distance $\mathit{
	d_{0}} = 1\mathit{m}$, $d_{k, b}^t$ is the geometric distance between the $k$th VU and the BS at slot $\mathit{t}$, and $\alpha_{k, b}$ is the corresponding path loss exponent. ${\bf{g}}_{k, b}^t$ is the frequency-dependent small-scale fading power component, which is assumed to be exponentially distributed with unit mean.
	
	Due to the link between BS and RIS is line-of-sight (LoS), the link between RIS and the vehicles is the same, therefore, these communication links undergo small-scale fading, which is modeled as the Rician fading with the pure LoS component \cite{r120, r121}. Since both the RIS and the BS are deployed at a fixed location, the RIS-BS link will remain static. Therefore, we can obtain the channel gain ${{\bf{h}}_{r,b}}$ between the RIS and BS as		
	\begin{equation}\label{eq4}
		{{\bf{h}}_{r,b}} = \sqrt {\rho {{\left( {{d_{r,b}}} \right)}^{ - {\alpha _{r,b}}}}} \sqrt {\frac{R}{{1 + R}}} {\bf{h}}_{r,b}^{LoS}.
	\end{equation}
	where ${d_{r,b}}$ is the geometric distance from the RIS to the BS, ${\alpha_{r,b}}$ is the path loss exponent of the RIS-BS link, and $R$ is the Rician coefficient associated with small-scale fading. LoS components ${\bf{h}}_{r,b}^{LoS}$ is defined as
	\begin{equation}\label{eq5}
		\begin{array}{r}
			{\bf{h}}_{r,b}^{LoS} = [1,{e^{ - j\frac{{2\pi }}{\lambda }{d_{r}}\sin \left( {{\theta _{r,b}}} \right)}}, \cdots ,
			{e^{ - j\frac{{2\pi }}{\lambda }\left( {N - 1} \right){d_{r}}\sin \left( {{\theta _{r,b}}} \right)}}{]^{\rm T}},
		\end{array}
	\end{equation}
	where $\lambda$ is the carrier wave length, $d_r$ is the interval between RIS elements, and $\theta _{r,b}$ is the departure angle of the signal from the RIS to the BS. Similarly, the channel gain ${\bf{h}}_{k,r}^t$ from the $k$th VU to the RIS at time slot $t$ is defined as
	\begin{equation}\label{eq6}
		{\bf{h}}_{k,r}^t = \sqrt {\rho {{\left( {d_{k,r}^t} \right)}^{ - {\alpha _{k,r}}}}} \sqrt {\frac{R}{{1 + R}}} {\bf{h}}_{k,r}^{t \; LoS},\forall k \in K,\forall t \in T,
	\end{equation}
	where, $d_{k,r}^t$ is the geometric distance between the $k$th vehicle and the RIS at time slot $t$, and ${\alpha _{k,r}}$ is the path loss exponent between the vehicle and the RIS. Note that ${\bf{h}}_{k,r}^{t \; LoS}$ is expressed as
	\begin{equation}\label{eq7}
		\begin{array}{c}
			{\bf{h}}_{k,r}^{t{\rm{ }}LoS} = [1,{e^{ - j\frac{{2\pi }}{\lambda }{d_r}\sin \left( {\theta _{k,r}^t} \right)}}, \cdots ,
			{e^{ - j\frac{{2\pi }}{\lambda }\left( {N - 1} \right){d_r}\sin \left( {\theta _{k,r}^t} \right)}}{]^{\rm T}},
		\end{array}
	\end{equation}
	where ${\theta _{k,r}^t}$ is the arrival angle of the signal from the $k$th vehicle to the RIS at time slot $t$.
	
	In this paper, multiple VUs in the VEC adopt the orthogonal frequency division multiplexing (OFDM) \cite{r1001} technique to communicate with the BS, and thus the interference over different subbands is not considered in the communication model. In addition, we consider direct communication links coexisting with RIS-assisted communication links. Thus, we can obtain the signal-to-noise ratio (SNR) between the $k$th VU and the BS at time slot $t$ as
	\begin{equation}\label{eq8}
		{\gamma _k}(t) = \frac{{P_{k,o}^t{{\left| {{{\left( {{{\mathbf{h}}_{r,b}}} \right)}^H}{\Theta ^t}{\mathbf{h}}_{k,r}^t + {\mathbf{h}}_{k,b}^t} \right|}^2}}}{{{\sigma ^2}}},\forall k \in \mathcal{K},\forall t \in \mathcal{T}
	\end{equation}
	where $P_{k,o}^t\in [0,{P_{\max,o}}]$ is the offloading power of $k$th VU at time slot $t$, and ${\sigma ^2}$ is thermal noise power. The diagonal phase-shift matrix of RIS is ${\Theta ^t} = diag[{\beta _1}{e^{j\theta _1^t}}, \cdots ,{\beta _n}{e^{j\theta _n^t}}, \cdots ,{\beta _N}{e^{j\theta _N^t}}],\forall n \in [1,N]$, ${\beta _n} \in \left[ {0,1} \right]$. Due to hardware constraint, phase-shift can only be selected from a finite discrete value set $\theta _n^t \in \Phi  = \left\{ {0,\frac{{2\pi }}{2^b}, \cdots ,\frac{{2\pi (2^b - 1)}}{2^b}} \right\}$, where $b$ controls the grid of phase-shift.
	
	When the $k$th VU chooses to offload tasks to the MEC server associated with the BS, based on the formula \eqref{eq7}, we can obtain the number of offloaded tasks processed by the $k$th VU at time slot $t$ as
	\begin{equation}\label{eq9}
		{q_{k,o}}(t) = \Delta t \times W{\log _2}(1 + {\gamma _k}(t)),
	\end{equation}
	where $W$ represents the sub-channel bandwidth.
	
	\subsection{Local Execution}
	When the $k$th VU selects to process tasks locally, we can obtain the size of tasks that can be processed locally at time slot $t$ as
	\begin{equation}\label{eq10}
		{q_{k,l}}(t) = \Delta t{f_k}(t)/L,
	\end{equation}
	where $L$ is denoted as the CPU frequency required to process one bit of task, 
	${f_k}(t) \in \left[ {0,{F_{\max }}} \right]$ is the CPU frequency scheduled by utilizing DVFS technology to adjust the chip voltage \cite{r122}, i.e., 
	\begin{equation}\label{eq11}
		{f_k}(t) = \sqrt[3]{{{p_{k,l}}(t)/c}},
	\end{equation}
	where  ${p_{k,l}}(t) \in [0,{P_{\max ,l}}]$ is the local execution power of the $k$th VU at time slot $t$, and $c$ is the effective selection capacitance.
	
	\subsection{Problem Formulation}
	To improve overall system efficiency, our goal is to address multiple performance metrics simultaneously. Specifically, the objective is to minimize the power consumption associated with task offloading and local processing, as well as the buffer length. Therefore, the multi-objective optimization problem for each the $k$th VU can be formulated as follows:
	\begin{subequations}\label{P1}
		\begin{equation}\label{eq12a}
			{P1}:\mathop {\min }\limits_{\theta _n^t,{p_{k,o}}(t),{p_{k,l}}(t)} \left\{ {\frac{1}{T}\sum\limits_{t = 1}^T {\left( {{p_{k,o}}(t) + {p_{k,l}}(t) + {q_k}(t)} \right)} } \right\},
		\end{equation}
		\begin{equation}\label{eq12b}
			{\rm{s}}{\rm{.t}}{\rm{.}}\quad0 < {p_{k,o}}(t) < {P_{\max ,o}},{\rm{ }}\forall k \in \mathcal{K},{\rm{ }}\forall t \in \mathcal{T},
		\end{equation}
		\begin{equation}\label{eq12c}
			0 < {p_{k,l}}(t) < {P_{\max ,l}},{\rm{ }}\forall k \in \mathcal{K},{\rm{ }}\forall t \in \mathcal{T},
		\end{equation}
		\begin{equation}\label{eq12d}
			{\rm{      }}\theta _n^t \in \Phi ,{\rm{            }}\forall n \in \mathcal{N},{\rm{ }}\forall t \in \mathcal{T},
		\end{equation}
	\end{subequations}
	where (\ref{eq12b}) and (\ref{eq12c}) represent the power constraints for the $k$th VU when offloading tasks and processing tasks locally, respectively. Due to the limitation of RIS hardware, the size of RIS phase-shift can only be selected within a limited range constrained by (\ref{eq12d}). Furthermore, the objective function is non-convex, so this optimization problem is difficult to be solved. To better address this issue, we propose a novel framework by combining the modified MADDPG algorithm with the centralized DDPG algorithm.  We will first use the DDPG algorithm to obtain the optimal coefficient of phase-shift, and then obtain the optimal power allocation scheme through the modified MADDPG algorithm.
	
	\section{RIS Phase-shift Optimization Based On DDPG}
	RIS is used as a tool to assist vehicle communication, and its goal is to maximize the information transmission rate of the vehicle by constantly adjusting the phase-shift coefficients to improve the performance of the system. In additional, \textit{P}1 involves optimizing the vehicle's offloading power, local execution power, and RIS phase-shift coefficients. Since the vehicle and RIS are different types of agents, so in order to solve \textit{P}1, we first solve the RIS phase-shift optimal problem during vehicle offloading by fixing the vehicle offloading power and then with the objective of maximizing the information transmission rate of the VU offloading link, we use the DDPG algorithm to train the phase-shift of the RIS, so as to obtain the optimal phase-shift coefficients of the RIS.
	Thus, the phase-shift coefficient optimization problem for RIS can be expressed as:
	\begin{subequations}\label{{P2}}
		\begin{equation}\label{eq13a}
			{P2}:\quad\mathop {\max }\limits_{\theta _n^t,} {\text{ }}{\log _2}(1 + {\gamma _k}(t)),
		\end{equation}
		\begin{equation}\label{eq13b}
			{\rm{s}}{\rm{.t}}{\rm{.}}\qquad{p_{k,o}}(t) = {P_{\max ,o}},{\rm{ }}\forall k \in \mathcal{K},{\rm{ }}\forall t \in \mathcal{T},
		\end{equation}
		\begin{equation}\label{eq13c}
			\quad{\rm{      }}\theta _n^t \in \Phi ,{\rm{            }}\forall n \in \mathcal{N},{\rm{ }}\forall t \in \mathcal{T}.
		\end{equation}
	\end{subequations}
	
	\subsection{DDPG for RIS Phase-shift Coefficients }
	In training the RIS phase shift coefficients with the DDPG algorithm, the BS acts as an agent interacting with the RIS-assisted vehicular network and adjusts the phase-shift coefficients of the RIS according to its strategy, aiming at solving the optimization problem $\mathit{P}2$. At time slot $t$, the BS obtains the current state $s_t$, and then obtains the corresponding action based on the current selected strategy, and transit to the next state $s_{t+1}$ and also obtains the reward $r_t$. this process can be formulated as 
	${e_t} = ({s_t},{a_t},{r_t},{s_{t + 1}})$. The relevant state space, action space, and reward function in this model are represented as follows:
	
	\textbf{State space:} the state obtained by the BS at time slot $t$ consists of the following components: phase-shift coefficient of the RIS $\theta _n^t$ , planar coordinate position of the $k$th VU $(x_k^t,y_k^t)$. In addition, through equation (\ref{eq8}), the SNR of VU $k$ at time slot $t$, i.e. ${\gamma _k}(t - 1)$ , depends on 
	${h_{r,b}}$ and $h_{k,r}^t$, which reflects the channel uncertainty of the VU at time slot $t$. Therefore, the state space of the agent is ${s_t} = [\theta _1^t, \cdots ,\theta _N^t,x_1^t,y_1^t, \cdots ,x_K^t,y_K^t,{\gamma _1}(t - 1),\cdots ,{\gamma _K}(t - 1)].$
	
	\textbf{Action space:} the BS acts as an agent whose action space is defined as ${a_t} = \{ \theta _1^t, \cdots ,\theta _n^t, \cdots ,\theta _N^t\} $ and $\theta _n^t$ is the phase-shift of the $n$th element of the RIS at time slot $t$. The BS maximizes the information transfer rate of the VU by adjusting the phase-shift of the RIS.
	
	\textbf{Reward function:} the reward function represents the objective to be optimized, in this subsection, we mainly optimize the information transmission rate of the VU by adjusting the RIS phase-shift coefficients and add a weighting factor $w$ for balance the reward, so the reward function is set as
	\begin{equation}\label{eq14}
		{r_t} = w{\log _2}(1 + {\gamma _k}(t))
	\end{equation}
	\begin{algorithm}[t]
		\caption{Training Stage of DDPG-based RIS Phase-shift Optimization}
		\label{al1}
		\KwIn{$\gamma$, $\tau$, $\theta_{\mu}$, $\theta_{Q}$}
		\KwOut{optimized $\theta_{\mu^*}$, $\theta_{Q^*}$}
		Randomly initialize the $\theta_{\mu}$, $\theta_{Q}$\;
		Initialize target networks by $\theta_{\mu'}\leftarrow\theta_{\mu}$, $\theta_{Q'}\leftarrow\theta_{Q}$\;
		Initialize replay experience buffer $\mathcal{D}$\;
		\For{each episode}
		{
			Reset simulation parameters for the system model;
			
			Receive initial observation state $s_{1}$;
			
			\For{each time step $t$ }
			{
				Generate RIS phase-shift coefficients according to the current policy and exploration noise $a_{t}=\mu (s;{\theta _\mu })+\Delta{n}$;
				
				Execute action $a_{t}$, observe reward $r_t$ and new state $s_t'$ from the system model;
				
				Store transition $(s_{t},a_{t},r_{t},s_{t}')$ in $\mathcal{D}$;
				
				\If {number of tuples in $\mathcal{D}$ is larger than $I$ }
				{
					Randomly sample a mini-batch of $I$ transitions tuples from $\mathcal{D}$;
					
					Update the critic network by minimizing the loss function according to Eq. (\ref{eq15});
					
					Update the actor network according to Eq. (\ref{eq17});
					
					Update target networks according to Eqs. (\ref{eq18}) and (\ref{eq19}).
				}
			}
		}
	\end{algorithm}
	
	\subsection{Concept of DDPG Algorithm}
	For high-dimensional state spaces, DDPG can be used to predict appropriate actions given the current state by deep neural networks that can learn complex features and patterns of the state. Therefore for the high-dimensional state and action space in this model, we use DDPG for training, and after obtaining the continuous action values, we can obtain the appropriate RIS phase-shift coefficients by scaling and rounding.
	In the DDPG algorithm, the state space and action space are usually represented by a deep neural network with two networks, actor and critic: the actor network is used to learn the strategy (i.e., action selection), and its goal is to maximize the estimation of the action value by the critic network, and the critic network is used to evaluate the value of the strategy, and learns to assess the evaluation by means of supervised learning merit of the actions output by the actor network. In addition, the DDPG algorithm uses a target actor network and a target critic network in order to ensure the stability of the algorithm. The network structure of these two networks is the same as that of the actor and critic networks, respectively.
	
	Firstly, initialize the parameters $\theta_{\mu}$ and $\theta _Q$ of actor network $\mu (s;{\theta _\mu })$ and critic network $Q(s,a;{\theta _Q})$, and set the parameters $\theta_{\mu'}$ and $\theta_{Q'}$ of target actor network $\mu'(s;\theta_{\mu'})$ and target critic network $Q'(s, a; \theta_{Q'})$, respectively. Then, at each time slot $t$, action $a_t$ is selected based on the current strategy $\mu$ and exploration noise $\Delta n$, while the reward and next state are observed by executing the action. Next, the experience of each step is stored in the experience buffer for random sampling training. At each time slot $t$, when the buffer length is greater than $I$, a batch of data is sampled from the experience replay buffer $\mathcal{D}$ which is used to update the parameters of the critic network to minimize the loss function
	\begin{equation}\label{eq15}
		\mathcal{L}({\theta _Q}) = \frac{1}{{I}}\sum\limits_i {({y_i} - Q(} {s_i},{a_i};{\theta _Q}){)^2},
	\end{equation}
	where the target value can be calculated as
	\begin{equation}\label{eq16}
		{y_i} = {r_i} + \gamma Q'({s_{i + 1}},\mu '({s_{i + 1}};{\theta _{\mu '}});{\theta _{Q'}}).
	\end{equation}
	
	At the same time, the actor network parameter $\theta_{\mu}$ is updated according to the gradient of the critic network
	\begin{equation}\label{eq17}
		{\nabla _{{\theta _\mu }}}J \approx \frac{1}{{I}}\sum\limits_i {{\nabla _a}} Q(s,a;{\theta _Q}){|_{s = {s_i},a = \mu ({s_i})}}\nabla {\theta _\mu }\mu (s;{\theta _\mu }){|_{{s_i}}},
	\end{equation}
	where $J$ is the target function of actor at the end of each time slot $t$, the target network parameters $\theta_{\mu'}$ and $\theta_{Q'}$ are softly updated to gradually approach the network parameters $\theta_{\mu}$ and $\theta _Q$ by the following parameter update process:
	\begin{equation}\label{eq18}
		\theta_{Q'} \leftarrow \tau \theta_{Q} + (1 - \tau) \theta_{Q'}
	\end{equation}
	\begin{equation}\label{eq19}
		\theta_{\mu'} \leftarrow \tau \theta_{\mu} + (1 - \tau) \theta_{\mu'}
	\end{equation}
	where $\tau$ is a soft update factor.
	
	Finally, after updating all the parameters, the above steps are repeated until a predetermined number of training episodes is reached or the stopping condition is satisfied.
	The related pseudo-code is shown in Algorithm \ref{al1}.
	
	\section{Power Allocation Optimization By Modified MADDPG Algorithm}
\begin{figure*}[t]
		\centering
		\includegraphics[width=6.7in, scale=1.00]{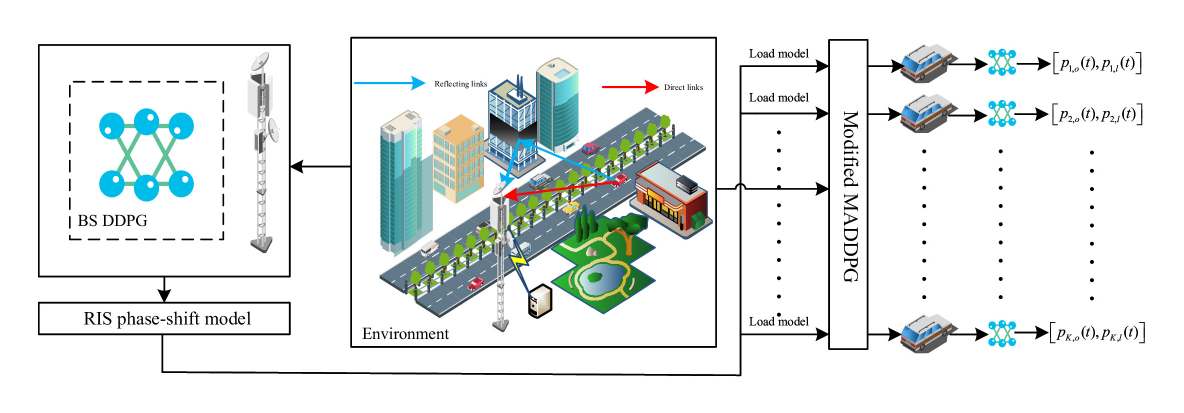}
		\caption{The proposed SARL-MARL framework. The framework contains a centralized DDPG model for RIS phase-shift optimization and a multi-agent model for power allocation}
		\label{fig2}
	\end{figure*}

	In this section, we will describe in detail the multi-agent environment and its associated states, actions and rewards, and finally, we will discuss the proposed MARL algorithm and its associated formulations. To be clear, our proposed framework for single-agent reinforcement learning and multi-agent reinforcement learning (SARL-MARL) is shown in Fig. \ref{fig2}, the relevant details are as follows.
	
	\subsection{Power Allocation Model for VEC}
	\begin{algorithm}[t]
		\caption{Modified MADDPG Power Allocation Algorithm with Joint RIS Phase-shift Model}\label{al2}
		Start environment simulator, generate vehicles and load RIS phase-shift model;\\
		Initialize global critic networks $Q_{{\psi _1}}^{{g_1}}$ and $Q_{{\psi _2}}^{{g_2}}$;\\
		Initialize target global critic networks $Q_{{\psi _1'}}^{{g_1}}$ and $Q_{{\psi _2'}}^{{g_2}}$;\\
		Initialize each agent's policy and critic networks;\\
		\For{each episode}
		{
			Reset or update simulation paramaters for the RIS-assisted VEC system;\\
			\For{each time step $t$}
			{
				Obtaining the corresponding RIS phase-shift coefficients through the RIS phase-shift model;\\
				\For{each agent $k$}
				{
					Observe $s_k^t$ and select action $a_k^t=\pi_\theta(s_k^t)$\\
					Receive local reward $r_{k,l}^t$;
				}
				${\bf{s}} = ({s_1},{s_2}, \cdots {s_K})$ and ${\bf{a}} = ({a_1},{a_2}, \cdots {a_K})$; \\
				Receive global reward $r_g^t$;\\
				Store $({{\bf{s}}^t},{{\bf{s}}^t},{\bf{r}}_l^t,r_g^t,{{\bf{s}}^{t + 1}})$ in replay buffer $\mathcal{D}$;\\
			
			\If{the size of the replay buffer is larger than $I$}
			{
				Randomly sample mini-batch of $I$ transitions tuples from $\mathcal{D}$;\\
				Update global critics by minimizing the loss according to Eq.(\ref{eq23});\\
				Update global target networks parameters: ${\psi _j'} \leftarrow \tau {\psi _j} + (1 - \tau ){\psi _j'}, j=1,2$;\\
				
				\If{episode mod $d$}
				{
					Train local critics and actors;\\
					\For{each agent $k$}
					{
						Update local critics by minimizing the loss according to Eq.($\ref{eq20}$);\\
						Update local actors according to Eq.($\ref{eq22}$);\\
						Update local target networks parameters: ${\theta _k'} \leftarrow \tau {\theta _k} + (1 - \tau ){\theta _k'}$, ${\phi _k'} \leftarrow \tau {\phi _k} + (1 - \tau ){\phi _k'}$;
					}
				}
			}
		}
		}	
	\end{algorithm}
	In the previous section, the BS was used as an agent to interact with the environment and an optimal model for the RIS phase-shift was trained based on Algorithm 1. In this section, we consider the VU as an agent, apply the multi-agent approach to interact with the environment, and obtain the power allocation decision through the corresponding policy. In each training step of power allocation, we will use the corresponding phase-shift coefficients based on the previous RIS phase-shift model, and based on this, we will obtain the channel of the VU, and thus determine the part of the training state of the VU to proceed to the next training step. The relevant state space, action space and reward function in the multi-agent model are defined as follows:
	
	\textbf{State space:} At time slot $t$, the state of each VU $k$ (agent $k$) consists of the following components: buffer length $q_k(t)$, the size of offloaded executed tasks $q_{k,o}(t)$, the size of locally executed tasks $q_{k,l}(t)$, and the offloaded and locally processed task overflows 
	${{q_{k,o}}(t) + {q_{k,l}}(t) - {q_k}(t)}$, the SNR of the $k$th VU at time slot $t$ ${\gamma _k}(t - 1)$. Therefore, the state space of the VU $k$ at slot $t$ can be formulated as ${s_k(t)} = \left[ {{q_k}(t),{q_{k,o}}(t),{q_{k,l}}(t),{q_{k,o}}(t) + {q_{k,l}}(t) - {q_k}(t),{\gamma _k}(t - 1)} \right].$
	
	\textbf{Action space:} as described above, the $k$th agent allocates the offloading power and the local power according to the corresponding policy, so that the action space of the $k$th agent at time slot $t$ is defined as 
	${a_k}(t) = \left[ {{p_{k,o}}(t),{p_{k,l}}(t)} \right]$.
	
	\textbf{Reward function:} in our proposed scheme, there are two aspects of rewards for each agent, one is the global reward reflecting the cooperation among agents, and the other is the local reward that is to help each agent to explore the optimal power allocation scheme.
	In DRL, rewards can be set flexibly, and a good reward can improve the performance of the system. Here, our goal is to optimize the agent $k$'s offloading and the local power level, as well as the corresponding buffer length. Thus the local and global rewards of the $k$th VU at time slot $t$ are
	\begin{equation}\label{eq20}
		{r_{k,l}} =  - \left[ {{w_1}({p_{k,o}}(t) + {p_{k,l}}(t)) + {w_2}{q_k}(t)} \right],
	\end{equation}
	and \begin{equation}\label{eq21}
		{r_{g}} = \frac{1}{K}\sum\limits_{k \in K} {{r_{k,l}}}.
	\end{equation}
	where $w_1$ and $w_2$ are the weighting factors to balance the relationship among the three different objectives. 
	
	\subsection{Multi-agent Deep Reinforcement Learning Algorithms}
	Based on DDPG algorithm, we propose a modified MADDPG algorithm that learns how to maximize both global and local rewards. It contains two main critics: a shared global critic and individual local critics. The global critic evaluates the states and actions of all agents collectively to optimize global rewards, while each agent has its own local critic that assesses rewards based on its specific local states and actions.
	Each agent operates with its own actor and critic networks for decision-making and action evaluation. Although agents can only observe their local environment, they can share experiences through a shared experience replay pool. During experience replay, agents store and sample from this shared pool of experiences, which enhances training efficiency and stability. This mechanism allows agents to learn from each other’s responses to the environment. The global reward signal promotes collaboration among agents by providing a unified reward feedback at each step, guiding the system towards optimized performance. In addition, we consider the impact of approximation errors in strategy and value updates on the global critic in MADDPG algorithm. Similar to \cite{r123}, to improve the MADDPG algorithm, we employ twin-delay deterministic strategy gradient to replace the global critic.
	
	Specifically, we consider a vehicular environment with $K$ vehicles (agents) and the policies for all agents are $\mathbf{\pi} = \{ {\pi _1},{\pi _2}, \cdots ,{\pi _K}\}$. The  $k$th agent's strategy $\pi_k$, local critic 
	${\mathop{\rm Q}\nolimits} _{{\phi _k}}^k$ and twin global critic 
	${\mathop{\rm Q}\nolimits} _{\psi 1}^{{g_1}}$, ${\mathop{\rm Q}\nolimits} _{\psi 2}^{{g_2}}$ are parameterized by ${\theta _k}$, $\phi_k$, ${\psi _1}$ and $\psi_2$, respectively. For each agent, the modified policy gradient can be written as
	\begin{equation}\label{eq22}
		\begin{aligned}
			\nabla J({\theta _k}) =& 
			\overbrace {{{\mathbb{E}}_{{\bf{s,a}}\sim \mathcal{D}}}\left[ {{\nabla _{{\theta _k}}}{\pi _k}({a_k}|{s_k}){\nabla _{{a_k}}}Q_{{\psi _j}}^{{g_j}}({\bf{s,a}})} \right]}^{Global\quad Critic} + \\& \underbrace {{{\mathbb{E}}_{{s_k},{a_k}\sim \mathcal{D}}}\left[ {{\nabla _{{\theta _k}}}{\pi _k}({a_k}|{s_k}){\nabla _{{a_k}}}Q_{{\phi _k}}^k({s_k},{a_k})} \right]}_{Local\quad Critic},
		\end{aligned}
	\end{equation}
	where ${\bf{s}} = ({s_1},{s_2}, \cdots {s_K})$ and 
	${\bf{a}} = ({a_1},{a_2}, \cdots {a_K})$ are the total state and action vectors, $\mathcal{D}$ is the replay buffer ,and ${a_k} = {\pi _k}({s_k})$ is the action which is chosen for agent $k$ according to its own policy $\pi_k$. Then the twin global critic $Q_{{\psi _j}}^{{g_j}}$ is updated to
	\begin{equation}\label{eq23}
		L({\psi _j}) = {{\rm E}_{{\bf{s,a,r,s'}}}}\left[ {{{(Q_{{\psi _j}}^{{g_j}}({\bf{s,a}}) - {y_g})}^2}} \right],{\rm{     }}j = 1,2,
	\end{equation}
	where 
	\begin{equation}\label{eq24}
		{y_g} = {r_g} + \gamma {\mathop {\min }\limits_{j = 1,2}}{\left. {Q_{\psi _j'}^{{g_j}}({\bf{s',a'}})} \right|_{a_k' = \pi _k'(s_k')}},
	\end{equation}
	where, ${\bf{\pi}}' = \{ \pi _1',\pi _2', \cdots ,\pi _K'\}$ is the target policy with parameter $\theta ' = \{ \theta _1',\theta _2', \cdots ,\theta _K'\}$. Similarly, the local critic $Q_{{\phi _k}}^k$ of the $k$th agent is updated to 
	\begin{equation}\label{eq25}
		{L^k}({\phi _k}) = {{\rm E}_{{s_k},{a_k},{r_k},s_k'}}\left[ {{{(Q_{{\phi _k}}^k({s_k},{a_k}) - y_l^k)}^2}} \right],
	\end{equation}
	where\begin{equation}\label{eq1}
		y_l^k = r_l^k + \gamma {\left. {Q_{\phi _k'}^k(s_k',a_k')} \right|_{a_k' = \pi _k'(s_k')}}.
	\end{equation}

	The detailed MARL algorithm is described in the Algorithm \ref{al2}. In addition, for the real-world implementation of Algorithm 2, it contains two aspects, on the one hand, for RIS phase-shift optimal coefficients, we need to obtain the vehicle's position, RIS phase-shift coefficients, etc. in each training step, and obtain the relevant RIS phase-shift optimal model after training. On the other hand, since we use the MADDPG algorithm to optimize the offloading power and the local execution power of the VU, this enables the VU to complete the resource allocation locally, and we only need to use the RIS phase-shift optimal coefficients allocation results from the BS in each training step, which greatly reduces the communication cost.
	
	\subsection{Testing Stage}
	\begin{algorithm}[t]
		\caption{Testing Stage for the proposed framework}
		\label{al3}
		\For{each test episode }
		{
			Reset simulation parameters for the VEC system model\;
			Obtaining RIS phase-shift optimization model correlation state $s_{RIS}^{1}$\;
			Obtaining power allocation model correlation state $s_{VU}^{1}$\;
			\For{each time step t }
			{
				Generate the RIS phase-shift coefficients according to the optimal policy $a_{RIS}=\mu(s_{RIS}^{1};\theta_\mu^*)$\;
				Calculate the information transmission rate based on the RIS phase-shift coefficient\;
				Generate the power for local process and computation offloading according to the optimal policy $a_{VU_k}=\pi_{\theta_k}(s_{VU_k}^{1};{\theta_k^*})$\;
				Execute action, observe reward and new state from the system model.
			}
		}
	\end{algorithm}
	The testing phase omits the updating of various network parameters and gradients, and the optimal phase-shift coefficients and power allocation scheme will be obtained from the trained model. We conduct tests based on the trained model, where each round contains 50 test slots, and each result is obtained by averaging the results of sequential 10 rounds. The pseudocode of the testing stage is shown in Algorithm \ref{al3}.
	
	\subsection{Complexity Analysis}
	In this section, we will analyze the complexity of the algorithm. The complexity is irrelevant since the RIS phase-shift model is obtained by training the DDPG algorithm and then loaded when training the VU power allocation model, rather than being trained simultaneously. Let $G_A$ and $G_C$ be the computational complexity of calculating the gradient for the actor network and the critic network, respectively, and $U_A$ and $U_C$ be the computational complexity of updating the parameters for the actor network and the critic network, respectively. Since the architecture of the target actor network and the target critic network is the same as the network structure of the two mentioned above, the associated computational gradient complexity and update parameter complexity are the same. Therefore, the complexity of the mentioned algorithms will be demonstrated as follows:
	\begin{itemize}
		\item[$\bullet$] \emph{DDPG:} 
		$\mathcal{O}\left( {(E \cdot S - I)({G_A} + {G_C} + 2{U_A} + 2{U_C})} \right)$
		
		\item[$\bullet$] \emph{TD3:} 
		$\mathcal{O}\left( {(E \cdot S - I)({G_A} + {G_C} + 3{U_A} + 3{U_C})} \right)$
		
		\item[$\bullet$] \emph{Modified MADDPG:}
		
		$
		\begin{gathered}
			\overbrace {\mathcal{O}\left( {K \cdot (E \cdot S - I)({G_A} + {G_C} + 2{U_A} + 2{U_C})} \right)}^{Local{\text{ }}agent} +  \hfill \\
			\underbrace {\mathcal{O}\left( {(E \cdot S - I)(2{G_C} + 4{U_C})} \right)}_{Global{\text{ }}critic} \hfill \\ 
		\end{gathered}$
	\end{itemize}
	where $E$ is the algorithm training episode, and each episode contains $S$ time slots for training, in addition, parameter updates and gradient computations are executed only when the tuples stored in the reply buffer is greater than $I$.
	
	\section{Simulation Results}
	\begin{table}[t]
		\caption{Values of the Parameters in the Experiments.}
		\label{tab1}
		\begin{center}
			\begin{tabular}{|c|c|c|c|}
				\hline
				\multicolumn{4}{|c|}{Parameters of System Model}\\
				\hline
				\textbf{Parameter}&{\textbf{Value}}&{\textbf{Parameter}}&{\textbf{Value}} \\
				\hline
				$K$ & 8 & $\eta$ & 3 Mbps \\
				\hline
				$N$ & 36 & $\sigma^2$ & -110 dBm \\
				\hline
				$b$ & 3 & $\alpha_{r,b}$ & 2.5 \\
				\hline
				$\alpha_{k,r}$ & 2.2 & $W$ & 1 MHz  \\
				\hline
				$L$ & 500 cycles/bit & $c$ & $10^{-28}$ \\
				\hline
				$F_{max}$ & 2.15 GHz & $P_{max,o}, P_{max,l}$ & 1 W\\
				\hline
				
				\hline
				\multicolumn{4}{|c|}{Parameters of DDPG and Modified MADDPG}\\
				\hline
				\textbf{Parameter} &\textbf{Value} &\textbf{Parameter} &\textbf{Value}\\
				\hline
				$\alpha_C$ &$0.001$ &$\alpha_A$ &$0.0001$\\
				\hline
				$\omega_{1}$ &$1$ &$\omega_{2}$ &$0.2$\\
				\hline
				$\omega$ &$1$&$d$ &$2$\\
				\hline
				$\gamma$ &$0.99$ &$\tau$ &$0.005$\\
				\hline
				$I$ &$64$ &$\mathcal{D}$ &$10^6$\\
				\hline
				$E$ &$1000$ &$S$ &$100$\\
				\hline
			\end{tabular}
		\end{center}
	\end{table}
	\begin{figure}[t]
		\centering
		\includegraphics[width=3.4in]{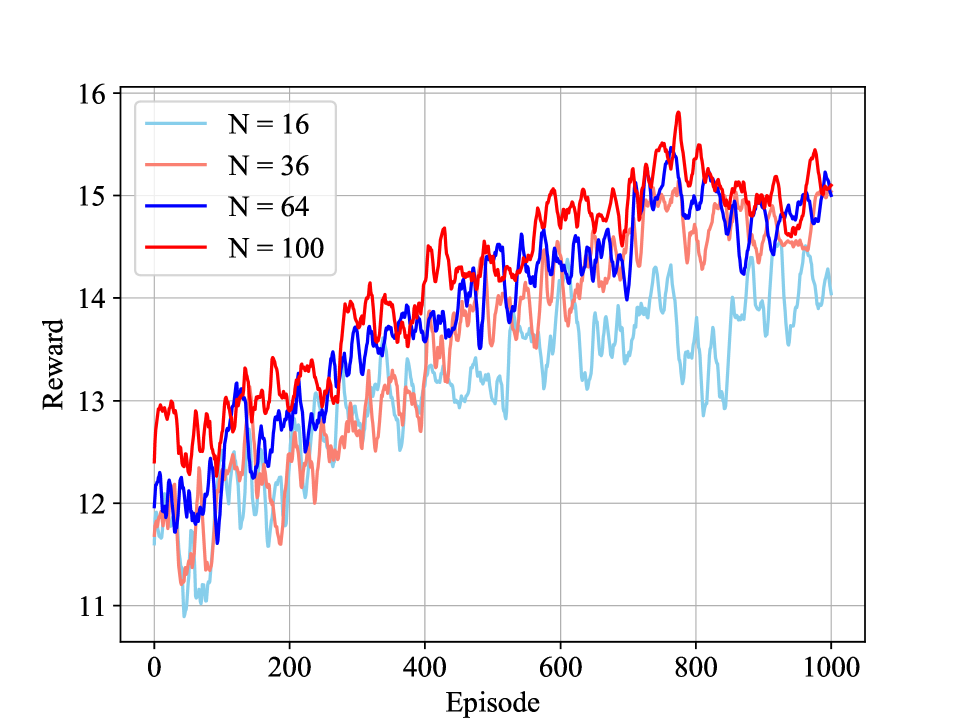}
		\caption{Comparison of rewards for RIS phase-shift training}
		\label{fig3}
	\end{figure}
	\begin{figure}[htbp]
		\centering
		\includegraphics[width=3.4in]{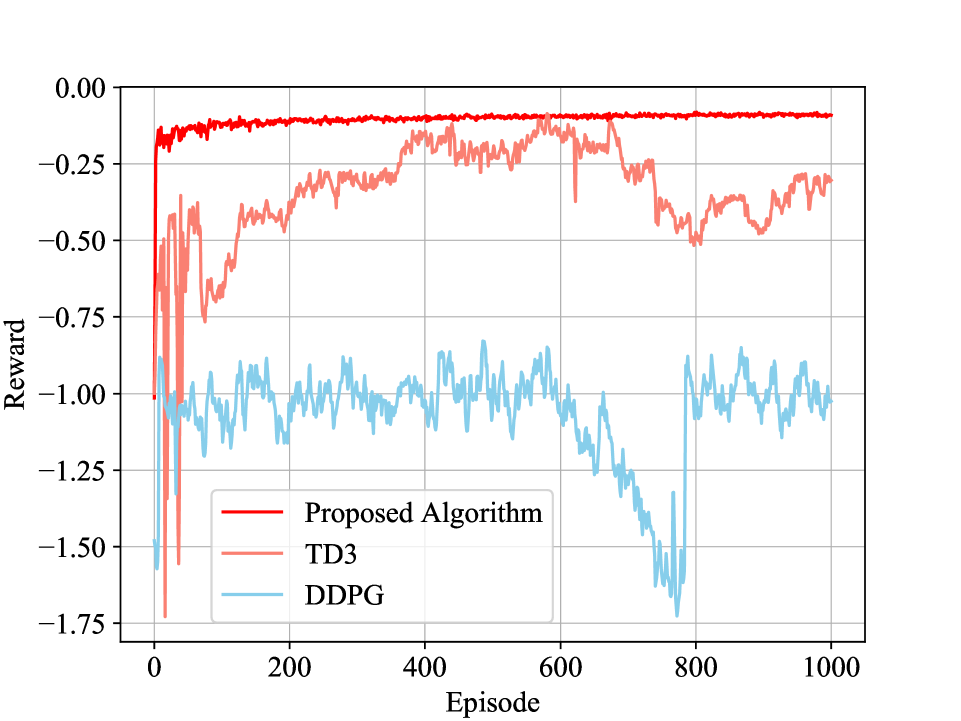}
		\caption{Reward convergence for VU power allocation training}
		\label{fig4}
	\end{figure}
	In this section, we perform simulation experiments to validate our proposed scheme. The simulation tool is Python 3.9. To verify the performance of the proposed scheme, we compare it with the traditional centralized DDPG and TD3 algorithm.
	It is assumed that the vehicle is travelling on an urban road and the position coordinates of the BS are $(0, 0, 25)$ and the position coordinates of the RIS are (250, 220, 25). $8$ vehicles are involved in the experiment, each with a randomly chosen speed within $10$ to $15$ m/s. The learning parameters of the DDPG scheme in Algorithm \ref{al1} and the modified MADDPG scheme in Algorithm \ref{al2} are same. The numbers of hidden layers for actor and critic networks are 2 and 3, respectively. The size of hidden layers for actor network is 512 and 256, and the size of hidden layers for critic network is 1024, 512 and 256. The key parameters are shown in Table \ref{tab1}.
	
	Fig. \ref{fig3} illustrates the different rewards between different numbers of RIS elements when optimizing the RIS phase-shift by the DDPG algorithm, and the higher the number of RIS elements, the higher the reward value obtained. According to the reward we set is related to the information transmission rate of the VU, which indirectly reflects the influence of the number of RIS elements on the information transmission rate of the VU, but in practice, we cannot blindly deploy more RIS elements, the more RIS elements, the higher the computational complexity brought about, and we have to consider comprehensively to achieve the optimal effect.
	
	\begin{figure*}[t]
		\centering
		\subfloat[]
		{\includegraphics[width=3.4in]{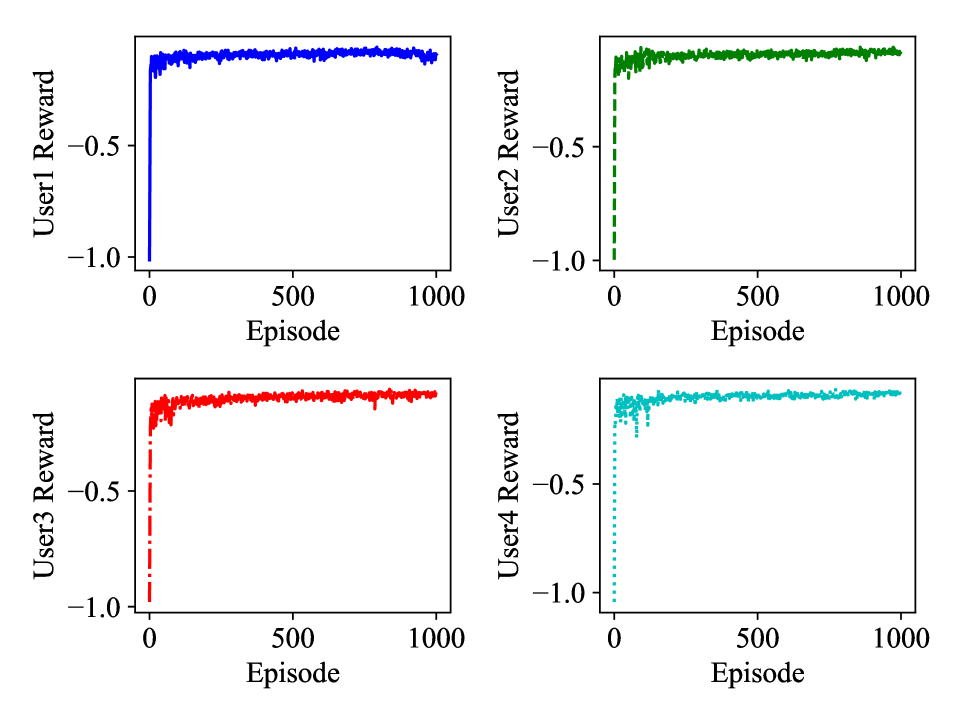}}
		\subfloat[]
		{\includegraphics[width=3.4in]{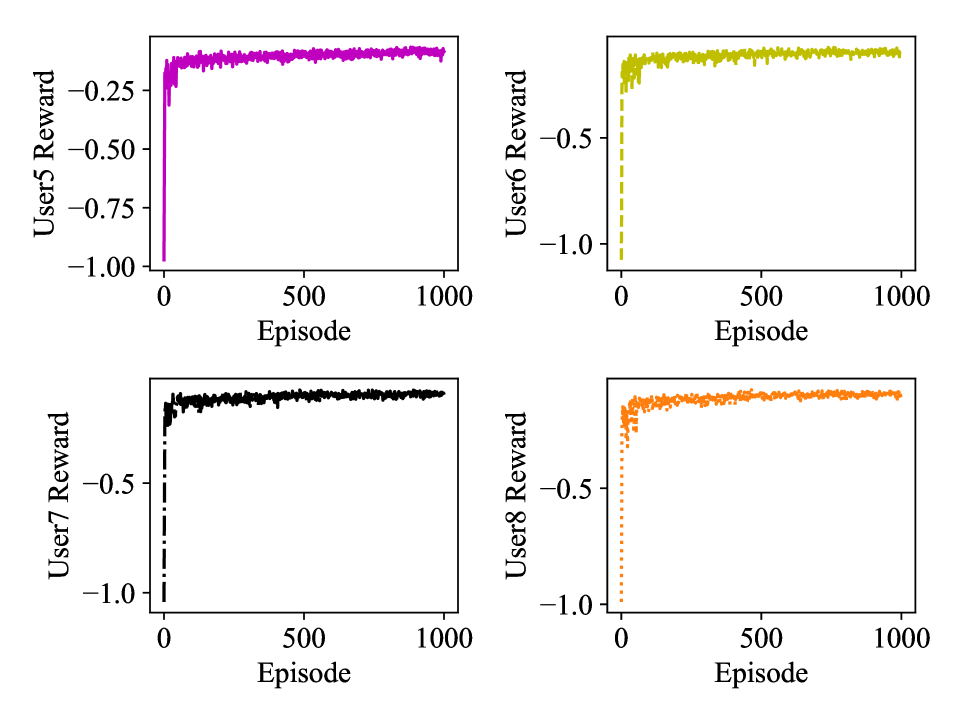}}
		\caption{Convergence of rewards for each user}
		\label{fig5}
	\end{figure*}
	
	Fig. \ref{fig4} presents the comparison between the proposed scheme and the TD3 and DDPG schemes in terms of reward convergence performance. The proposed scheme exhibits very stable convergence performance with less oscillations compared to the fully centralized TD3 and DDPG. TD3 and DDPG, as fully centralized approaches, require the observations and actions of all agents as inputs, which not only limits their cap ability to efficiently solve the individual performance problem. In such a way, it may lead to sub-optimal rewards and unstable convergence situations, and also puts a BS server imposes a significant information burden. On the contrary, the proposed framework combines the advantages of the SARL and MARL algorithms and exhibits superior convergence performance. It is capable of effectively selecting local and global reward maximization strategies for all agents, thus promoting cooperation among agents and further improving the overall system performance.
	
	\begin{figure}[t]
		\centering
		\includegraphics[width=3.4in]{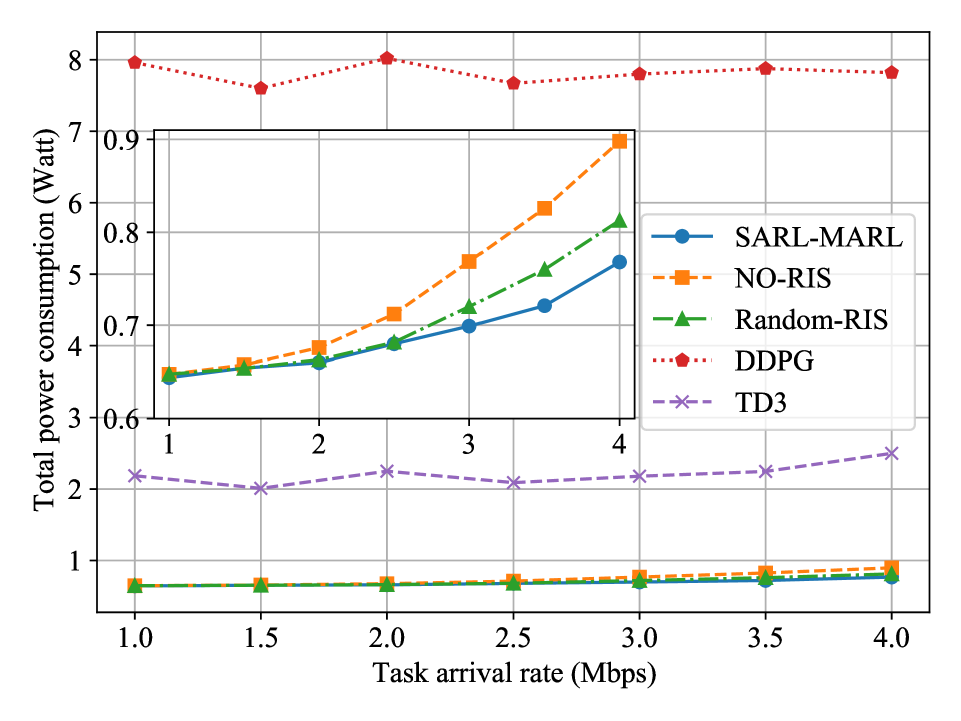}
		\caption{Total power consumption versus task arrival rate}
		\label{fig6}
	\end{figure}
	
	In addition, to further explore the advantages of Algorithm \ref{al2}, we observed the reward variations among different VUs, and from Fig. \ref{fig5}, we can observe that each agent exhibits a similar and stable reward curve, which is due to the fact that the phase shift coefficients of the RIS are invoked from the trained model, and the agents do not compete with each other, and all of them are aiming at reducing their own power consumption. However, it is unavoidable that the size or stability of the rewards may vary slightly due to the different positions or speeds of the agents in the environment. The key advantage of Algorithm \ref{al2} is that each VU chooses its actions based on its own state, and by setting the global critic, we ensure that the reward difference between VUs is not too large and can be kept stable. This design not only improves the overall collaborative efficiency of the system, but also ensures that each agent is able to make optimal decisions based on its unique conditions and goals in the collaborative effort.
	
	\begin{figure}[t]
		\centering
		\includegraphics[width=3.4in]{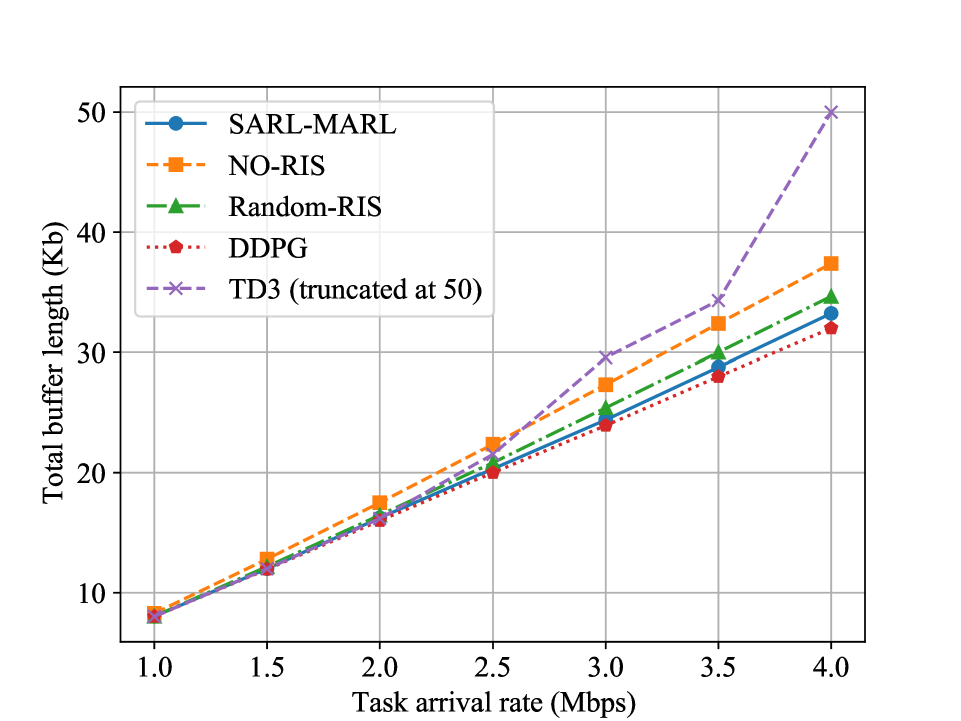}
		\caption{Buffer length versus task arrival rate}
		\label{fig7}
	\end{figure}
	Fig. \ref{fig6} illustrates the variation of power consumption of each VU at different task arrival rates. It can be observed that the power consumption of our proposed scheme increases with the task arrival rate. In contrast, the schemes without RIS assistance and with random RIS phase-shift have their power consumption slightly higher, highlighting the advantage of the DDPG algorithm for training the RIS phase-shifting coefficients. Finally, due to the relative advantage of the modified MADDPG algorithm, all these schemes consume significantly less power than the DDPG and TD3 schemes. Since DDPG and TD3 are fully centralized methods, they may fall into sub-optimal solutions and thus fail to achieve a good power allocation scheme. When the task arrival rate is increased from 1 Mbps to 4 Mbps, our proposed SARL-MARL scheme improves 19.25$\%$ in terms of power consumption. It is also to note that at a task arrival rate of 4 Mbps, our scheme reduces power consumption by 14.48$\%$ compared to the NO-RIS configuration and 90.18$\%$ compared to the fully centralized DDPG scheme.
	
	Fig. \ref{fig7} shows the variation of buffer lengths for each VU at different task arrival rates. As the task arrival rate increases, the buffer lengths of all VUs also increase. In particular, the buffer length of the DDPG algorithm is relatively low because it consumes too much power to transmit buffer tasks. On the contrary, the TD3 algorithm fails to balance power consumption and buffer length effectively, leading to task stacking at larger task arrival rates. At time slot $t$, when the task arrival rate is 4 M/bps, the buffer length of the TD3 algorithm is truncated to 50 Kb per ms, but the actual buffer length pileup reaches 750 Kb per ms. Our proposed scheme exhibits a relatively small buffer length with minimal power consumption, successfully balancing the relationship between power and buffer length. When the task arrival rate is 4 Mbps, the buffer length of our scheme decreases by 11.1$\%$ compared to the NO-RIS configuration while compared to the fully centralized DDPG scheme, it really increase 3.84$\%$ as the cost.
	
	\begin{figure}[t]
		\centering
		\includegraphics[width=3.4in]{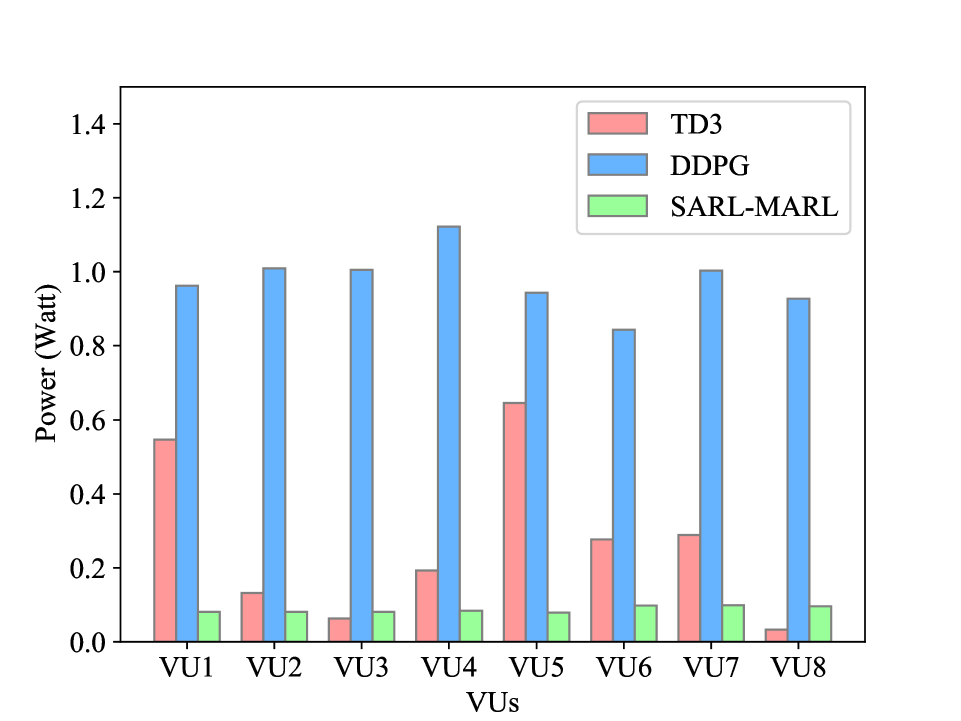}
		\caption{Per-user power consumption versus task arrival rate}
		\label{fig8}
	\end{figure}
	\begin{figure}[t]
		\centering
		\includegraphics[width=3.4in]{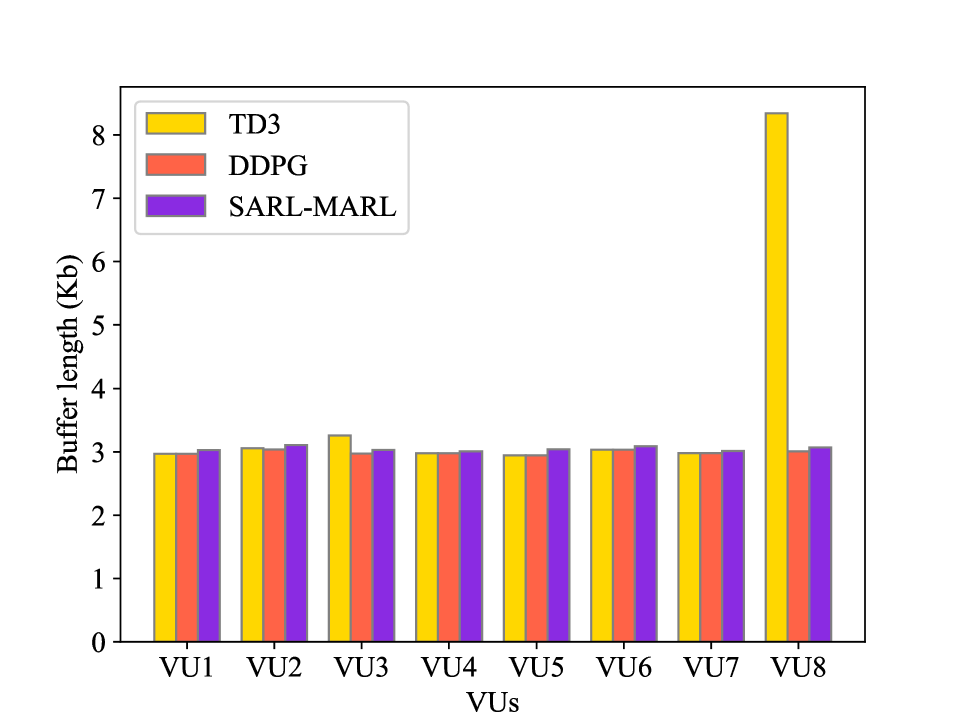}
		\caption{Per-user buffer length versus task arrival rate}
		\label{fig9}
	\end{figure}
	Finally, in order to study the impact of each algorithm on different VUs in more depth, we have analyze the power consumption versus buffer length for a task arrival rate of 3 M/bps. As can be seen in Figs. \ref{fig8} and \ref{fig9}, our proposed SARL-MARL scheme performs well in terms of power consumption and buffer length. The power consumption of each VU is kept at a low level, which is a significant advantage over the DDPG and TD3 schemes. In addition, in terms of buffer length, the TD3 scheme fails to effectively balance the buffer length among the VUs, resulting in less power being allocated to the 8th VU, and thus tasks cannot be transmitted in time and pile up. In contrast, under the DDPG algorithm, the buffer length of each VU is less, which is due to the excessive power allocated to each VU.
	
	\section{Conclusion}
	In this paper, the primary objective is to minimize both the power consumption and buffer length of VU. To address this challenging non-convex optimization problem, we decomposed it into two sub-problems. Firstly, a centralized DDPG algorithm was utilized to optimize the RIS phase-shift coefficients. Secondly, a modified MADDPG algorithm was employed to tackle the VU power allocation issue.
	The proposed framework has been evaluated through comprehensive simulations, demonstrating its superiority over alternative methods such as DDPG, TD3, and random phase-shift coefficient schemes in terms of reducing both power consumption and buffer length. These results underscore the effectiveness of leveraging advanced RL techniques to optimize RIS-enabled VU networks, enhancing their overall performance significantly. The conclusions are summarized as follows:
	\begin{itemize}
		\item [\textbullet] Our approach eases the computational pressure on the BS, and is able to cope with the complexity effectively, ensuring an efficient and stable convergence of the training process.
		
		\item [\textbullet]  We decomposed the problem into two sub-problems, which effectively reduced the computational complexity and enabled each agent to co-operate with each other to learn the optimal policy.
		
		\item [\textbullet] The deployment of RIS and the optimization of the RIS phase-shift coefficients positively impacted the performance of the whole system, and as the number of RIS increased, the VU information transmission rate is improved, highlighting its potential benefits.
	\end{itemize}

\end{document}